\documentclass[10pt,twocolumn,letterpaper]{article}

\usepackage[final]{cvpr}      
\usepackage{multirow}

\usepackage{pifont}
\usepackage{placeins} 
\newsavebox{\sggComponentsTableBox}
\newlength{\sggAblationTableWidth}

\definecolor{cvprblue}{rgb}{0.21,0.49,0.74}
\usepackage[pagebackref,breaklinks,colorlinks,allcolors=cvprblue]{hyperref}

\def\confName{CVPR}
\def\confYear{2027}

\title{SGG: Semantically Grounded Generation for Referring Video Object Segmentation}

\author{
{Fu Rong}$^{1}$\thanks{These authors contributed equally to this work.},
{Fengyuan Liu}$^{1}$\protect\footnotemark[1],
{Meng Lan}$^{2}$,
{Cong Yang}$^{1}$,
{Qian Zhang}$^{3}$,
{Lefei Zhang}$^{1}$\thanks{Corresponding author.} \\
$^{1}$School of Computer Science, Wuhan University \\
$^{2}$Hong Kong University of Science and Technology \quad
$^{3}$Horizon Robotics
}

\begin{document}
\maketitle
\begin{abstract}
Referring video object segmentation (RVOS) aims to identify and segment target objects throughout a video according to a natural-language expression. Pretrained video generative models offer rich visual-linguistic correspondences and spatiotemporal representations for RVOS. However, adapting these models poses two key challenges: distinguishing language-specified targets within reconstruction-oriented video latents and recovering accurate and complete object masks from spatially imperfect generated representations. To address these challenges, we propose Semantically Grounded Generation (\textbf{SGG}), a generative RVOS framework that leverages semantic representations to unify language-conditioned target generation and visually grounded mask refinement. Specifically, we introduce Semantically Enhanced Generation (SEG), which couples a trainable semantic expert operating on frozen DINOv3 features with the mask-generating expert via gated spatiotemporal attention. Together with text modulation and text cross-attention, SEG enables semantically grounded video-to-mask generation. Visually Refined Decoding then reformulates latent decoding as prompt-conditioned segmentation using SAM2. Latent-derived spatial features guide the aggregation of DINOv3 features into a shared object representation, which is projected into a sparse prompt. This representation further guides dynamic local aggregation of DINOv3 features to refine latent-derived spatial features for dense prompt construction. SAM2 integrates these complementary prompts with its own video features to produce accurate and complete target masks while preserving the referring semantics encoded in the generated latents. Extensive experiments on multiple RVOS benchmarks demonstrate that SGG achieves state-of-the-art performance.

\end{abstract}

\section{Introduction}
\label{sec:intro}

Referring video object segmentation (RVOS) aims to identify and segment target objects throughout a video according to a natural-language expression~\cite{Referformer,he2024decoupling,lan2024bidirectional,liang2026m2}. Beyond appearance-based tracking, RVOS requires interpreting linguistic cues about attributes, actions, and relationships to distinguish the referred objects from other instances. Effective RVOS therefore hinges on both language-conditioned target understanding and precise spatial delineation despite variations in appearance and motion~\cite{Referformer,he2024decoupling}.

\begin{figure}[t]
    \centering
    \includegraphics[width=\linewidth]{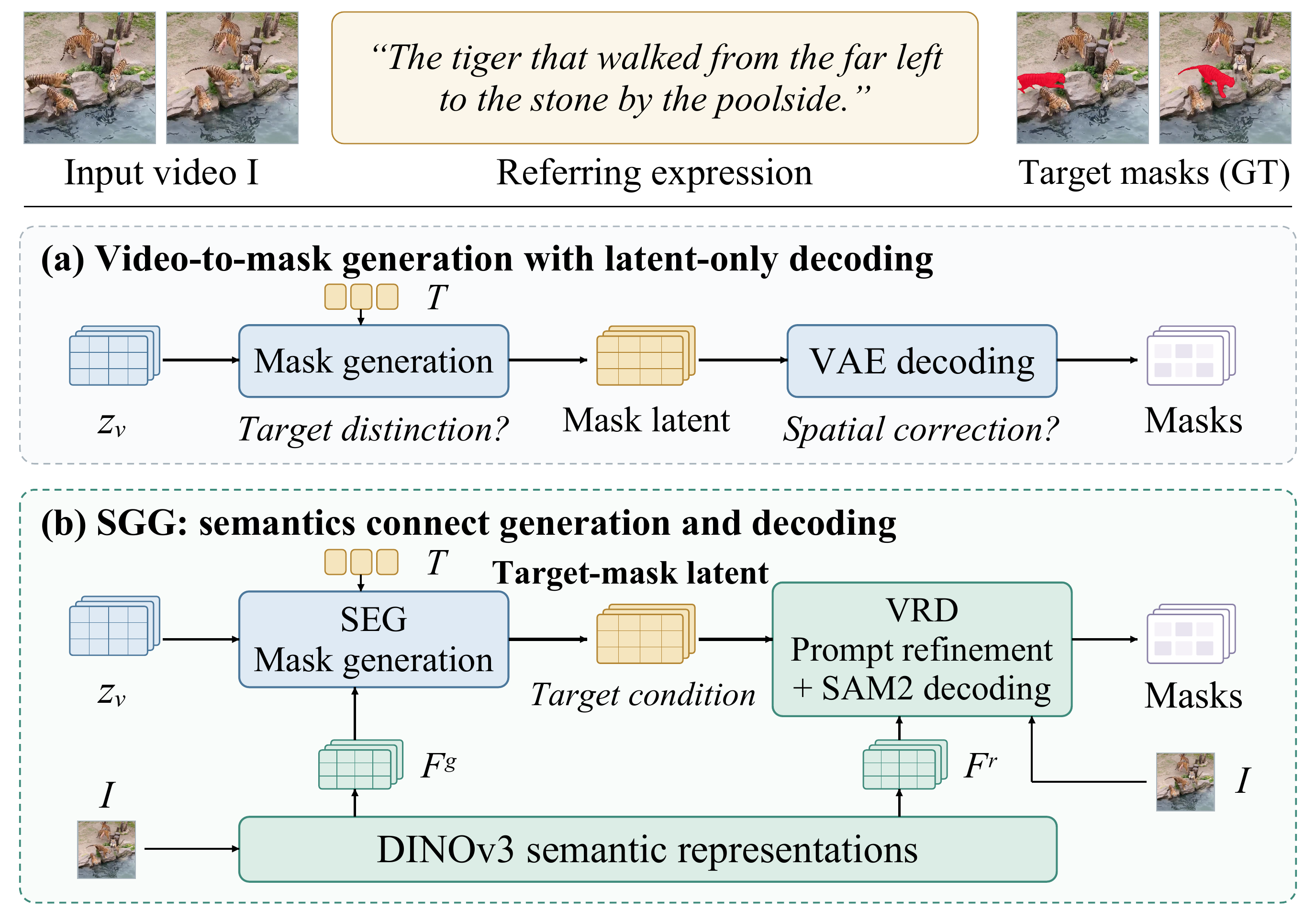}
    \caption{Generative RVOS pipelines: (a) a VAE-based baseline; (b) our SGG. SGG integrates DINOv3 semantic guidance into target-mask generation and prompt-conditioned decoding, improving target identification and the spatial accuracy and completeness of predicted masks.}
    \label{fig1}
\end{figure}

Recent RVOS methods~\cite{Referformer,rong2025mpgsam2,cuttano2025samwise,zhang2026streamingrvos} have advanced these capabilities through language-conditioned object queries, the adaptation of promptable segmentation models~\cite{ravi2024sam,rong2026rs2}, and the integration of multimodal large language models~\cite{yuan2025sa2va,gong2025devil}. Meanwhile, pretrained text-to-video models offer an alternative foundation for RVOS, as learning to synthesize videos from text endows them with visual-linguistic correspondences and spatiotemporal representations that can be transferred to segmentation. Along this direction, VD-IT~\cite{zhu2024exploring} transfers features from a frozen video diffusion model to a segmentation decoder, while FlowRVS learns a language-conditioned flow from video latents to target-mask latents~\cite{wang2026flowrvs}. Despite these advances, effectively leveraging video generative priors for RVOS remains challenging in two fundamental aspects.

First, generative RVOS requires distinguishing the language-specified target from the surrounding scene. Video VAE latents are optimized for appearance reconstruction and thus jointly encode target objects, non-target instances, and background content~\cite{wang2026stwam}. However, faithful scene reconstruction and language-conditioned target identification impose different representational requirements: the former emphasizes appearance preservation, whereas the latter requires object-level semantic discrimination. Although existing generative models incorporate text conditioning, this alone does not explicitly disentangle target-specific semantics from scene-level representations. The video-to-mask transformation must therefore learn to distinguish the referred object under linguistic guidance while preserving its spatiotemporal structure.

Second, accurate target identification does not necessarily translate into spatially precise and complete segmentation masks. Generated masks may still exhibit fragmented regions, missing object parts, or spurious foreground responses. This limitation is particularly evident in VAE-based decoding, where masks are reconstructed from generated target-mask latents without direct access to visual features of the observed video frames. Consequently, spatial inaccuracies in the generated latents may propagate into the decoded masks, as the decoder lacks direct visual evidence to correct erroneous predictions or recover missing object regions. Achieving accurate and complete segmentation therefore requires grounding the decoding process in the observed visual content rather than relying solely on generated representations.

To address these challenges, we propose \textbf{SGG}, a \textbf{S}emantically \textbf{G}rounded \textbf{G}eneration framework for RVOS that leverages DINOv3 semantic representations to unify language-conditioned target-mask generation and visually grounded decoding. As illustrated in \cref{fig1}, the two stages are connected through a generated target-mask latent, which encodes the referring semantics and is subsequently refined using visual features from the observed video. Specifically, Semantically Enhanced Generation (\textbf{SEG}) introduces a trainable semantic expert operating on frozen DINOv3 features and couples it with the mask-generating expert via gated spatiotemporal attention~\cite{simeoni2026dinov3}. Text Modulation (TM) and text cross-attention jointly guide the video-to-mask transformation according to the referring expression. To encourage target-specific semantic modeling, the semantic expert is supervised with DINOv3 features extracted from rendered target masks, guiding its learned representations to capture the spatial extent of the referred target.

To translate the generated target-mask latent into spatially accurate and complete segmentation masks, we further introduce Visually Refined Decoding (\textbf{VRD}), which reformulates latent decoding as prompt-conditioned segmentation using SAM2. By leveraging both generated target representations and visual features from the observed video, VRD provides direct visual evidence for refining spatially imperfect latent representations. Specifically, latent-derived spatial features guide the aggregation of DINOv3 features into a shared object representation, which is projected into a sparse prompt to convey object-level foreground information to SAM2. The object representation further guides dynamic local aggregation of DINOv3 features via object-to-frame matching to refine the latent-derived spatial features for dense prompt construction. SAM2 integrates these complementary prompts with its own video-frame features to produce spatially precise and complete target masks~\cite{ravi2024sam}.

Extensive experiments on multiple RVOS benchmarks demonstrate that SGG achieves state-of-the-art segmentation performance. Our main contributions are summarized as follows:

\begin{itemize}
\item We propose \textbf{SGG}, a generative RVOS framework that unifies language-conditioned target-mask generation and visually grounded decoding through DINOv3 semantic representations, with the generated target-mask latent serving as the interface between the two stages.
\item We introduce Semantically Enhanced Generation, which integrates DINOv3 semantic representations into video-to-mask flow modeling through gated spatiotemporal expert interaction and text modulation to facilitate discriminative target representation.
\item We develop visually refined decoding, which reformulates latent decoding as prompt-conditioned segmentation with SAM2. A shared object representation enables object-level sparse prompting and guides the refinement of latent-derived spatial features using DINOv3 features for dense prompt construction, facilitating accurate and complete target-mask prediction.
\end{itemize}

\begin{figure*}[t]
    \centering
    \includegraphics[width=\linewidth]{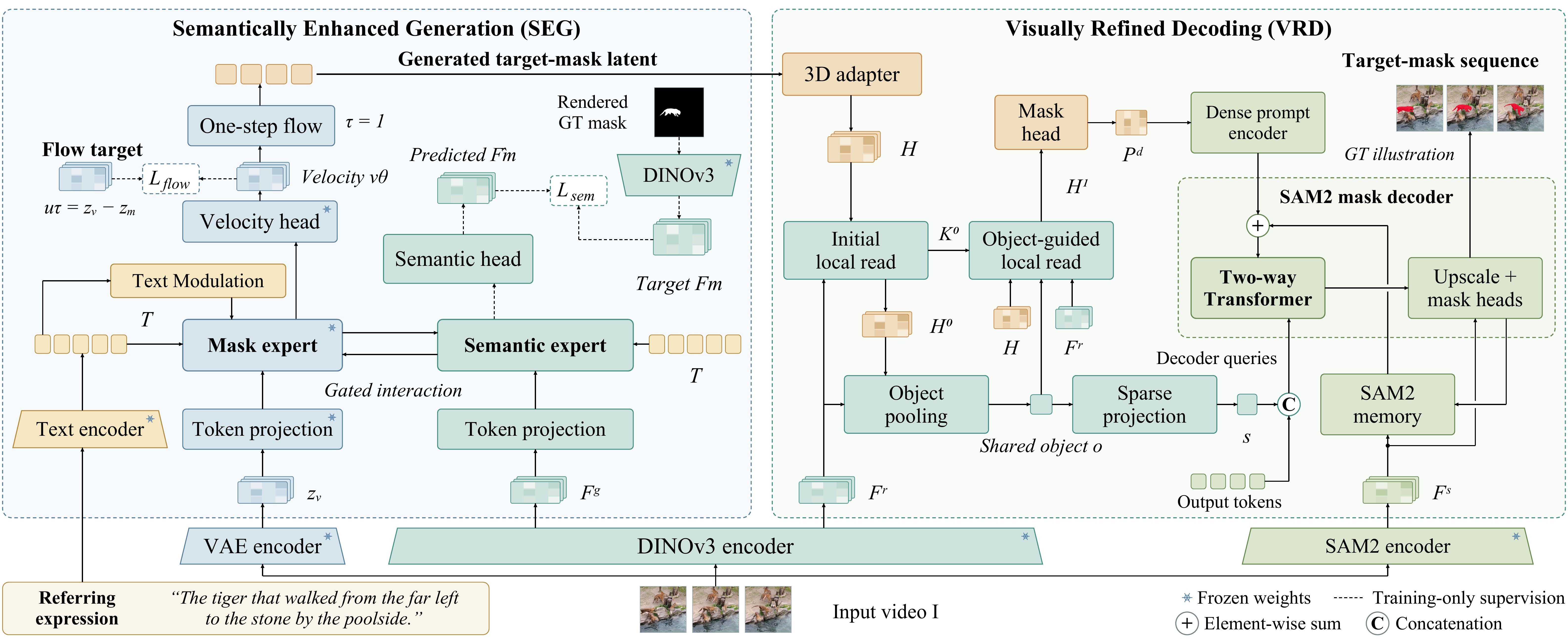}
    \caption{Overview of SGG. Semantically Enhanced Generation integrates DINOv3 semantics into language-conditioned target-mask generation through gated spatiotemporal expert interaction and text modulation. DINOv3 features of rendered target masks supervise the semantic expert. Visually Refined Decoding uses the generated target-mask latent and DINOv3 video-frame features to aggregate a shared object representation. The object-guided local read uses object-to-frame matching to guide local aggregation and refine latent-derived features for dense prompting, while sparse projection maps the object representation to a sparse prompt. SAM2 combines both prompts with its video-frame features to predict target masks.}
    \label{framework}
\end{figure*}

\section{Related work}
\noindent\textbf{Referring Video Object Segmentation.}
RVOS aims to segment objects specified by natural-language expressions throughout a video. Early Transformer-based approaches primarily rely on object queries to model cross-modal interactions and temporal dependencies. ReferFormer~\cite{Referformer} introduces language queries to identify and segment referred objects across frames. Subsequent studies~\cite{miao2023spectrum,luo2024soc,yuan2024losh,he2024decoupling,liang2025referdino} further enhance cross-modal alignment, temporal representation learning, and visual grounding~\cite{liu2024grounding,lan2024language}. Among them, DsHmp~\cite{he2024decoupling} decouples static and motion cues in referring expressions to guide segmentation and temporal reasoning, respectively. ReferDINO~\cite{liang2025referdino} extends the region-level vision--language alignment of GroundingDINO~\cite{liu2024grounding} to video segmentation, leveraging predicted bounding boxes to guide deformable mask decoding.

More recently, the emergence of promptable segmentation foundation models has motivated their adaptation to RVOS through language-guided prompting. RefSAM~\cite{li2024refsamefficientlyadaptingsegmenting} adapts SAM by introducing cross-modal prompt generation and an implicit tracking token to incorporate linguistic and temporal information. Building upon the SAM2, SAMWISE~\cite{cuttano2025samwise} introduces linguistic and temporal adapters for language-conditioned video segmentation, while MPG-SAM2~\cite{rong2025mpgsam2} leverages mask priors and global context to construct complementary dense and sparse prompts.

In parallel, multimodal large language models (MLLMs) have been employed to enhance language reasoning. VideoLISA~\cite{bai2024videolisa} combines sparse-dense video sampling with a single tracking token to enable cross-frame segmentation. VideoGLaMM~\cite{munasinghe2025videoglamm} introduces visual-to-language and language-to-visual adapters to connect video representations with an LLM and translate its outputs into mask-decoding inputs. To better integrate global semantic reasoning with local temporal tracking, GLUS~\cite{lin2025glus} combines sparse context frames with consecutive query frames and jointly trains an MLLM with segmentation memory. StreamingRVOS~\cite{zhang2026streamingrvos} further explores efficient streaming segmentation by feeding previous segmentation embeddings back into the MLLM to preserve temporal context.

\noindent\textbf{Generative Models.}
Latent diffusion models~\cite{rombach2022latent} perform denoising in a compressed representation space, substantially reducing the computational cost of high-resolution synthesis, while diffusion Transformers~\cite{peebles2023dit} provide a scalable architecture for latent generative modeling. Extending these advances to the video domain, models such as CogVideoX~\cite{yang2025cogvideox} and Wan~\cite{wan2025} combine spatiotemporal compression with text-conditioned Transformers to jointly model visual appearance and temporal dynamics. Recent studies further improve the efficiency and temporal consistency of long-video generation~\cite{yin2025causvid}. Self Forcing~\cite{huang2025selfforcing} trains autoregressive models on their own generated context to mitigate the training--inference discrepancy. LongLive~\cite{yang2026longlive} introduces streaming long-video training and prompt-aware cache updates, whereas CausalCine~\cite{meng2026causalcine} extends causal generation to multi-shot videos through memory retrieval.

Beyond diffusion-based formulations, flow matching~\cite{lipman2023flow} provides a general framework for learning continuous transformations between probability distributions by modeling time-dependent velocity fields along probability paths. Rectified Flow~\cite{liu2023rectifiedflow} further improves transport efficiency by straightening trajectories through reflow. Beyond visual synthesis, these generative formulations and their learned visual representations have also shown potential for dense prediction. DepthFM~\cite{gui2025depthfm} transfers pretrained generative knowledge to image-to-depth transport, while SemFlow~\cite{wang2024semflow} formulates semantic segmentation and image synthesis as bidirectional transport between image and semantic-mask distributions.

These advances motivate the exploration of pretrained video generative models for RVOS, where their learned visual--linguistic correspondences and spatiotemporal representations can facilitate language-guided mask prediction. VD-IT~\cite{zhu2024exploring} adopts a feature-transfer paradigm, extracting representations from a frozen text-to-video diffusion model for subsequent segmentation. In contrast, FlowRVS~\cite{wang2026flowrvs} directly formulates RVOS as a generative process by learning a language-conditioned flow from video latents to target-mask latents. Building upon this generative paradigm, our work integrates semantic guidance into target-mask generation and grounds latent decoding in the observed visual content, addressing both target-specific semantic discrimination and spatially accurate mask prediction.

\section{Method}
\label{sec:method}

\subsection{Overview}
\label{sec:sgg-method-1}

Given an \(N\)-frame video \(\mathcal I=\{I_i\}_{i=1}^{N}\) and a referring expression \(\mathcal E\), SGG connects language-conditioned target generation with visual refinement through a target-mask latent interface to predict \(\mathcal M=\{M_i\}_{i=1}^{N}\). SEG integrates DINOv3 semantics and text conditioning into video-to-mask flow modeling. At inference, the generated target-mask latent guides VRD to refine spatial representations and aggregate object representations from DINOv3 features. The resulting dense and sparse prompts condition SAM2 segmentation using features from its own video-frame encoder. The generation and decoding stages are trained separately.

\subsection{Feature Extraction}
\label{sec:sgg-method-2}

We use frozen Wan VAE~\cite{wan2025}, UMT5~\cite{chung2023unimax}, and DINOv3 encoders~\cite{simeoni2026dinov3}. The video latent \(\mathbf z_v\) is normalized using the VAE's channel statistics, with posterior sampling during training and the posterior mean at inference. UMT5 produces text embeddings \(\mathbf T\). DINOv3 extracts per-frame semantic features \(\mathbf F_i^g\) and \(\mathbf F_i^r\) from half-resolution and full-resolution inputs for generation and refinement, respectively; \(\mathbf F^g=\{\mathbf F_i^g\}_{i=1}^N\) denotes the generation feature sequence. A frozen SAM2 encoder independently extracts multi-scale video-frame features \(\mathbf F_i^s\) for decoding~\cite{ravi2024sam}.

\subsection{Semantically Enhanced Generation}
\label{sec:sgg-method-3}

\begin{figure}[t]
  \centering
  \includegraphics[width=\linewidth]{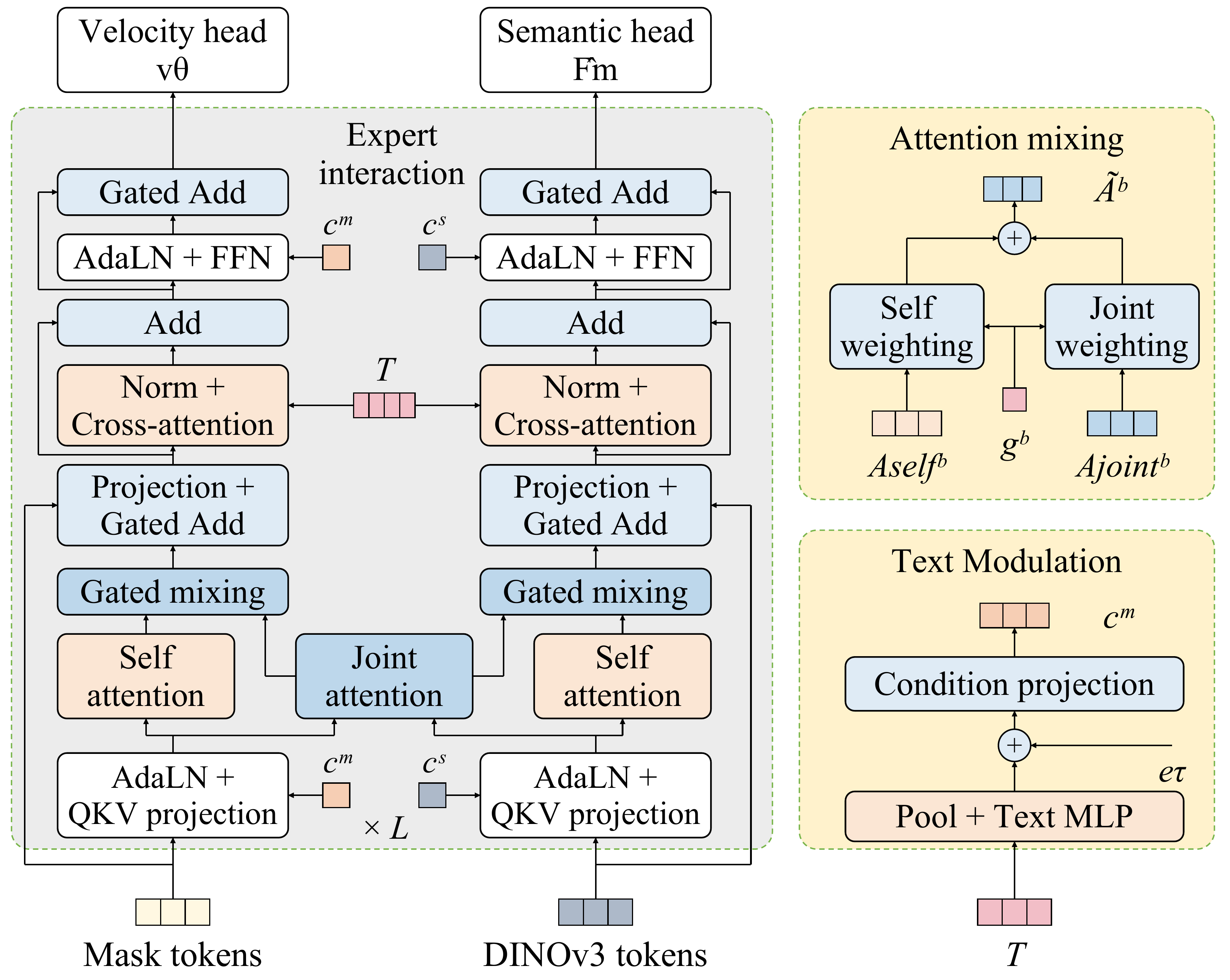}
  \caption {Structure of the semantically enhanced generation module.}
  \label{fig:seg_detail}
\end{figure}

Reconstruction-oriented video latents preserve appearance across the scene, whereas the output mask selects the object described by language. SEG introduces complementary DINOv3 semantics to support this target distinction within video-to-mask flow modeling. Gated expert interaction integrates the two representations, and text modulation works with text cross-attention to guide the mask expert toward the referred object. The detailed structure is shown in \cref{fig:seg_detail}.

\noindent\textbf{Video-to-Mask Flow}. We formulate target prediction through flow matching~\cite{wang2026flowrvs}. The target-mask latent \(\mathbf z_m\) is the VAE posterior mean of binary masks replicated across three channels and mapped to \([-1,1]\), without channel normalization. With video at \(\tau=1\) and masks at \(\tau=0\), the training path and single-step inference are:
\begin{equation}
\label{eq:sgg-flow-inference}
\begin{aligned}
\mathbf z_\tau&=\tau\mathbf z_v+(1-\tau)\mathbf z_m,
\quad\mathbf u_\tau=\mathbf z_v-\mathbf z_m,\\
\widehat{\mathbf z}_m
&=\mathbf z_v-v_\theta(\mathbf z_v,1;\mathbf T,\mathbf F^g).
\end{aligned}
\end{equation}
Here, \(v_\theta\) predicts the path velocity, and \(\widehat{\mathbf z}_m\) is passed directly to VRD.

\noindent\textbf{Gated Expert Interaction.} Separate projections tokenize \(\mathbf z_\tau\) and \(\mathbf F^g\) for the mask and semantic experts, which comprise \(L\) paired transformer blocks. Their queries and keys include branch-specific normalization and rotary coordinates aligned across resolutions~\cite{wang2026stwam}. At each layer and head, branch \(b\in\{m,s\}\) combines self-attention with attention over both experts:
\begin{equation}
\label{eq:sgg-expert-attention}
\begin{aligned}
A_{\mathrm{self}}^b&=\operatorname{Attn}(Q^b,K^b,V^b),\\
A_{\mathrm{joint}}^b&=\operatorname{Attn}
\left(Q^b,[K^m;K^s],[V^m;V^s]\right),\\
\widetilde A^b&=(1-g^b)A_{\mathrm{self}}^b+g^b A_{\mathrm{joint}}^b.
\end{aligned}
\end{equation}
Here, \(\operatorname{Attn}(Q,K,V)=\operatorname{softmax}(QK^\top/\sqrt{d_h})V\), with head dimension \(d_h\); layer and head indices are omitted for clarity. The learned gates \(g^b\) are clipped to \([0,1]\) independently for each layer and head. Joint attention exchanges spatial and temporal information bidirectionally; each expert retains its output projection, residual updates, text cross-attention, and feed-forward layers.

\noindent\textbf{Text Modulation.} To condition the mask expert's latent updates at the sentence level, we average the non-padding embeddings in \(\mathbf T\) into \(\overline{\mathbf t}\) and modulate its time embedding:
\begin{equation}
\label{eq:sgg-text-modulation}
\widetilde{\mathbf e}_\tau
=\mathbf e_\tau+\operatorname{MLP}_{\mathrm{text}}(\overline{\mathbf t}).
\end{equation}

The resulting embedding, projected to the block condition \(\mathbf c^m\), controls normalization shifts, scales, and residual gates within the mask expert's transformer blocks, guiding latent updates alongside token-level text cross-attention. The mask output head retains the original time embedding, while \(\mathbf c^s\) denotes the semantic expert's independently projected time condition.

\subsection{Visually Refined Decoding}
\label{sec:sgg-method-4}

\begin{figure}[t]
  \centering
  \includegraphics[width=\linewidth]{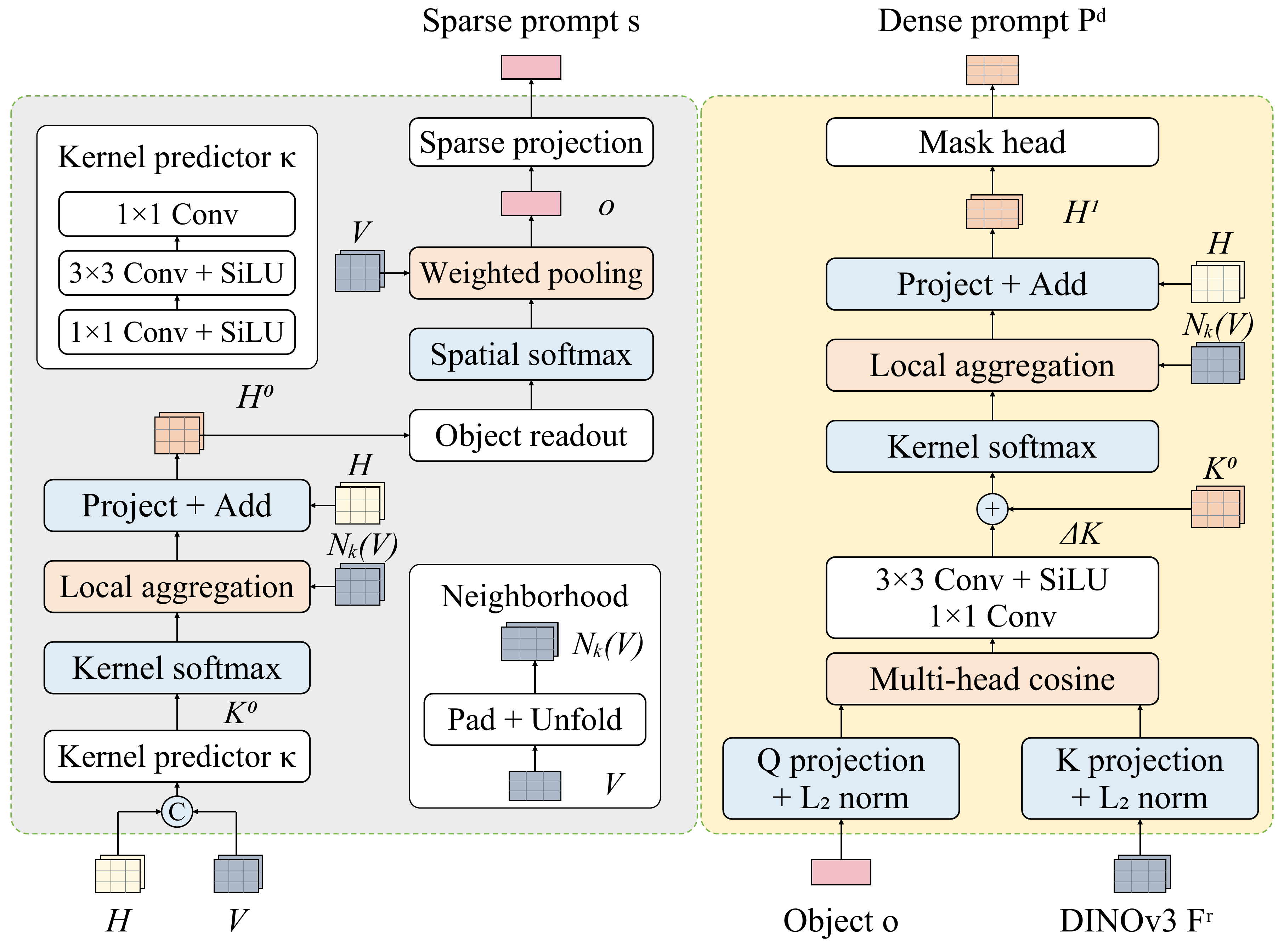}
  \caption{Structure of the visually refined decoding module.}
  \label{fig:vrd_detail}
\end{figure}

Spatial errors in the generated target-mask latent can compromise the accuracy and completeness of decoded masks. VRD uses DINOv3 features to refine latent-derived features and aggregate a shared object representation, whose projection provides a sparse prompt for SAM2. Object-to-frame matching guides dynamic local refinement for dense prompt construction. The detailed structure is shown in \cref{fig:vrd_detail}.

A 3D adapter and temporal interpolation convert the target-mask latent into per-frame features \(H_i\). Omitting the frame index, let \(V=\phi_v(\mathbf F^r)\) be projected DINOv3 features, and let \(\rho_D,\rho_H\) resize features to the DINOv3 and adapter grids. For kernel logits \(K\), local aggregation is
\begin{equation}
\label{eq:sgg-local-aggregation}
\mathcal A(V,K)(p)
=\sum_{\delta\in\mathcal N_k}
\operatorname{softmax}_{\delta}(K(p,:))\,V(p+\delta),
\end{equation}
where \(\mathcal N_k\) is a \(k\times k\) neighborhood and weights are normalized over its offsets and shared across channels. The notation \(\mathcal N_k(V)\) in the module diagram denotes the corresponding neighborhood features extracted from \(V\). The initial local read in \cref{framework} predicts logits \(K^0=\kappa([\rho_D(H);V])\) and uses them to update \(H\) to \(H^0\). Object pooling then derives the shared object representation:
\begin{equation}
\label{eq:sgg-object-pooling}
\begin{aligned}
H^0&=H+\rho_H\!\left(\phi_u(\mathcal A(V,K^0))\right),\\
a(p)&=\operatorname{softmax}_{p}\!\left(\phi_o(\rho_D(H^0))(p)\right),\\
\mathbf o&=\sum_p a(p)V(p).
\end{aligned}
\end{equation}
Here, \(\phi_u\) projects the local update and \(\phi_o\) predicts spatial pooling logits. Object pooling in Eq.~\eqref{eq:sgg-object-pooling} expands into Object readout, Spatial softmax, and Weighted pooling in \cref{fig:vrd_detail}: \(H^0\) determines the spatial weights \(a(p)\), which are applied to \(V\) to obtain \(\mathbf o\). The pooled object representation is projected into a sparse prompt for SAM2. We also match \(\mathbf o\) against the original DINOv3 features in an \(A\)-head space. The object projection includes LayerNorm before splitting into heads. With head-wise projected queries \(\phi_q^a(\mathbf o)\) and keys \(\phi_k^a(\mathbf F^r(p))\), object-to-frame cosine relations adjust the local kernels:
\begin{equation}
\label{eq:sgg-object-refinement}
\begin{aligned}
r_a(p)&=\operatorname{cos}\!\left(
\phi_q^a(\mathbf o),\phi_k^a(\mathbf F^r(p))\right),\\
\Delta K&=\psi_\Delta([r_1,\ldots,r_A]),\\
H^1&=H+\rho_H\!\left(\phi_u(\mathcal A(V,K^0+\Delta K))\right).
\end{aligned}
\end{equation}

The object-guided local read in \cref{framework} follows Eq.~\eqref{eq:sgg-object-refinement}, with cosine similarity computed after channel-wise normalization within each head. The convolutional predictor \(\psi_\Delta\) converts matching relations into kernel-logit corrections. Both reads share \(V\) and \(\phi_u\), with residuals added to \(H\). The refined features and the object vector from Eq.~\eqref{eq:sgg-object-pooling} yield complementary dense and sparse prompts, respectively:
\begin{equation}
\label{eq:sgg-prompts}
P^d=\rho_P(\phi_d(H^1)),
\qquad
\mathbf s=\phi_s(\mathbf o).
\end{equation}
Here, \(\phi_d\) predicts single-channel mask logits, \(\rho_P\) resizes and bounds them, and \(\phi_s\) produces one sparse prompt. Thus, the shared object representation supports both local spatial refinement and sparse prompt construction.

\subsection{Prompt Encoding and Mask Decoding}
\label{sec:sgg-method-5}

Mask decoding is performed by SAM2, conditioned on the dense and sparse prompts constructed from the target-mask latent and DINOv3 features~\cite{ravi2024sam}. Its prompt encoder maps the continuous dense prompt \(P_i^d\) to dense embeddings; the sparse prompt \(\mathbf s_i\) is appended to its sparse embeddings. Dense embeddings are added to video-frame features extracted by the SAM2 encoder, while sparse embeddings join the learned output tokens as decoder queries. SAM2 combines these inputs with higher-resolution features and memory to predict masks, which are resized to the original resolution. At inference, the generated target-mask latent conveys the referring condition to prompt construction. The dense prompt supplies refined spatial guidance, while the sparse prompt summarizes foreground appearance for the decoder.

\subsection{Training Objective}
\label{sec:sgg-method-6}

\noindent\textbf{Generation Stage.} SEG regresses the velocity in Eq.~\eqref{eq:sgg-flow-inference}:
\begin{equation}
\label{eq:sgg-flow-loss}
\mathcal L_{\mathrm{flow}}
=\mathbb E\left[
N_z^{-1}\left\|
v_\theta(\mathbf z_\tau,\tau;\mathbf T,\mathbf F^g)-\mathbf u_\tau
\right\|_F^2
\right],
\end{equation}
where \(N_z\) is the latent element count and the expectation is over training triplets and sampled times. Time sampling mixes the video endpoint with quantized uniform samples, emphasizing the initialization used for single-step inference. The semantic expert additionally predicts DINOv3 features \(\widehat F_m\) of fixed-color rendered target masks, supervised by frozen features \(F_m\). This objective encourages the semantic expert to learn representations that reflect the spatial extent of the referred target. Let \(e_p=\|\widehat F_m(p)-F_m(p)\|_2^2/C_D\), where \(C_D\) is the DINOv3 dimension, and let \(w_p^{\mathrm{fg}}\) be mask occupancy pooled to the feature grid, with \(w_p^{\mathrm{bg}}=1-w_p^{\mathrm{fg}}\). Averaging over nonempty regions \(\mathcal A_M\subseteq\{\mathrm{fg},\mathrm{bg}\}\), we use
\begin{equation}
\label{eq:sgg-semantic-loss}
\mathcal L_{\mathrm{sem}}
=\mathbb E\left[
\frac{1}{|\mathcal A_M|}
\sum_{r\in\mathcal A_M}
\frac{\sum_p w_p^r e_p}{\sum_p w_p^r}
\right].
\end{equation}

The index \(p\) spans the clip's semantic feature grid; region balancing is performed per clip before averaging. With the mask expert frozen, we train the semantic expert, its projections, the interaction gates, and the text modulation network.

\noindent\textbf{Decoding stage.} VRD is trained on the ground-truth target-mask latent \(\mathbf z_m\) and video-frame features, without running the generator. Final predictions receive focal and soft Dice supervision~\cite{lin2017focal,milletari2016v}. The stage-specific objectives are
\begin{equation}
\label{eq:sgg-training-objectives}
\begin{aligned}
\mathcal L_{\mathrm{SEG}}
&=\lambda_{\mathrm{flow}}\mathcal L_{\mathrm{flow}}+\lambda_{\mathrm{sem}}\mathcal L_{\mathrm{sem}},\\
\mathcal L_{\mathrm{VRD}}
&=\lambda_{\mathrm{focal}}\mathcal L_{\mathrm{focal}}+\lambda_{\mathrm{Dice}}\mathcal L_{\mathrm{Dice}},
\end{aligned}
\end{equation}
where \(\lambda_{\mathrm{flow}}\), \(\lambda_{\mathrm{sem}}\), \(\lambda_{\mathrm{focal}}\), and \(\lambda_{\mathrm{Dice}}\) weight the flow-matching, semantic, focal, and Dice losses, respectively. Focal loss is averaged over pixels and frames, while Dice is computed per frame and then averaged. We freeze the VAE, DINOv3, and SAM2 video-frame encoders, and optimize the latent adapter, refinement and prompt modules, and SAM2's prompt encoder, mask decoder, and memory modules. At inference, the generated target-mask latent replaces its ground-truth counterpart.

\section{Experiments}
\label{sec:experiments}

\subsection{Datasets and Metrics}

\noindent\textbf{Datasets.} Our evaluation covers Ref-YouTube-VOS~\cite{Urvos}, MeViS~\cite{ding2023mevis}, and Ref-DAVIS17~\cite{khoreva2018video}, which provide complementary tests of appearance-based referring, motion understanding, and performance across video scenes. The Ref-YouTube-VOS training partition contains 3,471 videos paired with 12,913 expressions, with a further 202 videos reserved for validation. MeViS provides 28,570 expressions over 2,006 videos and emphasizes object identification through motion. Ref-DAVIS17 adds referring expressions to objects in 90 DAVIS17 videos.

\noindent\textbf{Evaluation Metrics.} Segmentation quality is measured by region similarity $\mathcal{J}$ and boundary accuracy $\mathcal{F}$, together with their mean $\mathcal{J}\&\mathcal{F}$; scores and absolute score differences are expressed in \%. We submit Ref-YouTube-VOS and MeViS valid predictions to their official evaluation servers, as the validation masks are not released. On MeViS, comparisons with state-of-the-art methods use the valid split, whereas ablation studies use valid\_u. For Ref-DAVIS17, we follow the official language-referring protocol and evaluate the Ref-YouTube-VOS-trained model without additional DAVIS training or ground-truth dense prompts.

\subsection{Implementation Details}

\noindent\textbf{Model Settings.} SGG combines a Wan2.1-T2V-1.3B~\cite{wan2025} target-mask latent generator with semantic features from DINOv3 ViT-L/16~\cite{simeoni2026dinov3} and a SAM2-Hiera-Large~\cite{ravi2024sam} segmentation decoder. The mask-generating expert starts from FlowRVS~\cite{wang2026flowrvs} parameters; the Wan VAE and UMT5 encode the video and text, respectively. We use an inference resolution of $480\times832$ and resize frames to $1024\times1024$ for SAM2. Within SEG, DINOv3 receives half-resolution frames, with spatial and temporal feature strides set to 32 and 1. VRD instead extracts full-resolution DINOv3 features at spatial stride 16. Text modulation uses a hidden dimension of 256. The local refinement neighborhood is $3\times3$, while object-to-frame matching operates in a 256-dimensional space with eight heads. One sparse prompt is formed from the aggregated object representation for each frame.

\noindent\textbf{Training Details.} Generation and decoding are trained in separate stages, each using AdamW, BF16 mixed precision, and one 17-frame clip per GPU. For MeViS, we initialize SEG from a FlowRVS model pretrained on MeViS, optimize it for two epochs, and then train VRD for two epochs. Learning rates are assigned by component: $6\times10^{-5}$ for SEG, $5\times10^{-5}$ for the latent adapter and VRD prompt modules, and $2\times10^{-6}$ for trainable SAM2 components. Weight decay is $5\times10^{-4}$. We keep the VAE, UMT5, DINOv3, SAM2 frame encoder, and original mask-generating expert fixed. The loss weights are set to $\lambda_{\mathrm{flow}}=1$, $\lambda_{\mathrm{sem}}=5$, and $\lambda_{\mathrm{focal}}=\lambda_{\mathrm{Dice}}=1$. VRD receives the VAE posterior means of ground-truth masks and applies focal and Dice losses to the final predictions, allowing it to be trained without executing the generator. For Ref-YouTube-VOS, both stages use its training set; the resulting model is also evaluated on Ref-DAVIS17 with no further fine-tuning.

\subsection{Comparison with State-of-the-Art Methods}

\noindent\textbf{Ref-YouTube-VOS and Ref-DAVIS17.} \cref{Refytvos_RefDavis} summarizes the performance of SGG and state-of-the-art RVOS methods on both benchmarks. SGG reaches 72.7\% $\mathcal{J}\&\mathcal{F}$ on Ref-YouTube-VOS, surpassing the generative methods VD-IT~\cite{zhu2024exploring} and FlowRVS~\cite{wang2026flowrvs} by 6.2\% and 3.1\%, respectively. Compared with the recently published methods GeoLaV~\cite{zhu2026geolav} and DeRVOS~\cite{cheng2026dervos}, SGG improves $\mathcal{J}\&\mathcal{F}$ by 2.2\% and 2.7\%, respectively. On Ref-DAVIS17, SGG achieves 75.5\% $\mathcal{J}\&\mathcal{F}$, exceeding the leading methods STAC~\cite{ariff2026stac}, FlowRVS, and GeoLaV by 1.9\%, 2.2\%, and 3.0\%, respectively.

\begin{table}[!t]
\centering
\footnotesize
\setlength{\tabcolsep}{0.8pt}
\caption{Quantitative results on Ref-YouTube-VOS and Ref-DAVIS17. Bold and underlined entries indicate the highest and second-highest scores in each column.}
\label{Refytvos_RefDavis}
\begin{tabular}{@{}l | c | c c c | c c c@{}}
\toprule
\multirow{2}{*}{Method} & \multirow{2}{*}{Reference} & \multicolumn{3}{c|}{Ref-YouTube-VOS} & \multicolumn{3}{c}{Ref-DAVIS17} \\
 & & $\mathcal{J}\&\mathcal{F}$ & $\mathcal{J}$ & $\mathcal{F}$ & $\mathcal{J}\&\mathcal{F}$ & $\mathcal{J}$ & $\mathcal{F}$ \\
\midrule
ReferFormer~\cite{Referformer} & CVPR'22 & 62.9 & 61.3 & 64.6 & 61.1 & 58.1 & 64.1 \\
SgMg~\cite{miao2023spectrum} & ICCV'23 & 65.7 & 63.9 & 67.4 & 63.3 & 60.6 & 66.0 \\
TempCD~\cite{tang2023temporal} & ICCV'23 & 65.8 & 63.6 & 68.0 & 64.6 & 61.6 & 67.6 \\
SOC~\cite{luo2024soc} & NeurIPS'23 & 66.0 & 64.1 & 67.9 & 64.2 & 61.0 & 67.4 \\
MUTR~\cite{yan2024referred} & AAAI'24 & 68.4 & 66.4 & 70.4 & 68.0 & 64.8 & 71.3 \\
LoSh~\cite{yuan2024losh} & CVPR'24 & 67.2 & 65.4 & 69.0 & 64.3 & 61.8 & 66.8 \\
DsHmp~\cite{he2024decoupling} & CVPR'24 & 67.1 & 65.0 & 69.1 & 64.9 & 61.7 & 68.1 \\
VISA~\cite{yan2024visa} & ECCV'24 & 63.0 & 61.4 & 64.7 & 70.4 & 67.0 & 73.8 \\
VD-IT~\cite{zhu2024exploring} & ECCV'24 & 66.5 & 64.4 & 68.5 & 69.4 & 66.2 & 72.6 \\
SSA~\cite{pan2025ssa} & CVPR'25 & 64.3 & 62.2 & 66.4 & 67.3 & 64.0 & 70.7 \\
DMVS~\cite{fang2025dmvs} & CVPR'25 & 64.3 & 62.4 & 66.2 & 65.2 & 62.2 & 68.2 \\
SAMWISE~\cite{cuttano2025samwise} & CVPR'25 & 69.2 & 67.8 & 70.6 & 70.6 & 67.4 & 74.5 \\
ReferDINO~\cite{liang2025referdino} & ICCV'25 & 69.3 & 67.0 & 71.5 & 68.9 & 65.1 & 72.9 \\
FlowRVS~\cite{wang2026flowrvs} & ICLR'26 & 69.6 & 67.1 & \underline{72.1} & 73.3 & 68.4 & \underline{78.2} \\
DeRVOS~\cite{cheng2026dervos} & CVPR'26 & 70.0 & 68.0 & 71.9 & 70.9 & 68.3 & 73.5 \\
STAC~\cite{ariff2026stac} & ECCV'26 & 65.2 & 63.5 & 66.9 & \underline{73.6} & \underline{70.1} & 77.2 \\
GeoLaV~\cite{zhu2026geolav} & ECCV'26 & \underline{70.5} & \underline{69.1} & 71.8 & 72.5 & 69.9 & 75.2 \\
\midrule
SGG (ours) & \textbf{---} & \textbf{72.7} & \textbf{70.4} & \textbf{75.0} & \textbf{75.5} & \textbf{70.7} & \textbf{80.3} \\
\bottomrule
\end{tabular}
\end{table}

\noindent\textbf{MeViS.} The MeViS valid comparison in \cref{Mevis} evaluates SGG against state-of-the-art RVOS methods on motion-based expressions, where object behavior over time is essential to target identification. SGG obtains 54.8\% $\mathcal{J}\&\mathcal{F}$, exceeding the leading methods in the table, DeRVOS~\cite{cheng2026dervos}, FlowRVS~\cite{wang2026flowrvs}, and GeoLaV~\cite{zhu2026geolav}, by 3.0\%, 3.7\%, and 4.8\%, respectively. This performance demonstrates its effectiveness in segmenting targets described through motion. Relative to the target-mask latent generation of FlowRVS, SGG incorporates semantic guidance during generation and uses the generated target representation to condition visual refinement at decoding.

\begin{table}[t]
\centering
\footnotesize
\setlength{\tabcolsep}{3pt}
\caption{Quantitative results on MeViS valid. Bold and underlined entries indicate the highest and second-highest scores in each column.}
\label{Mevis}
\begin{tabular}{l | c | c c c}
\toprule
Method & Reference & $\mathcal{J}\&\mathcal{F}$ & $\mathcal{J}$ & $\mathcal{F}$ \\
\midrule
URVOS~\cite{Urvos} & ECCV'20 & 27.8 & 25.7 & 29.9 \\
LBDT~\cite{ding2022language} & CVPR'22 & 29.3 & 27.8 & 30.8 \\
MTTR~\cite{botach2022end} & CVPR'22 & 30.0 & 28.8 & 31.2 \\
ReferFormer~\cite{Referformer} & CVPR'22 & 31.0 & 29.8 & 32.2 \\
VLT+TC~\cite{ding2021vision} & TPAMI'22 & 35.5 & 33.6 & 37.3 \\
LMPM~\cite{ding2023mevis} & ICCV'23 & 37.2 & 34.2 & 40.2 \\
DsHmp~\cite{he2024decoupling} & CVPR'24 & 46.4 & 43.0 & 49.8 \\
VISA~\cite{yan2024visa} & ECCV'24 & 44.5 & 41.8 & 47.1 \\
SSA~\cite{pan2025ssa} & CVPR'25 & 48.6 & 44.0 & 53.2 \\
DMVS~\cite{fang2025dmvs} & CVPR'25 & 48.6 & 44.2 & 52.9 \\
SAMWISE~\cite{cuttano2025samwise} & CVPR'25 & 49.5 & 46.6 & 52.4 \\
ReferDINO~\cite{liang2025referdino} & ICCV'25 & 49.3 & 44.7 & 53.9 \\
FlowRVS~\cite{wang2026flowrvs} & ICLR'26 & 51.1 & 47.6 & 54.6 \\
DeRVOS~\cite{cheng2026dervos} & CVPR'26 & \underline{51.8} & \underline{48.1} & \underline{55.4} \\
STAC~\cite{ariff2026stac} & ECCV'26 & 49.9 & 47.2 & 52.5 \\
GeoLaV~\cite{zhu2026geolav} & ECCV'26 & 50.0 & 47.4 & 52.9 \\

\midrule
SGG (ours) & \textbf{---} & \textbf{54.8} & \textbf{51.7} & \textbf{57.9} \\
\bottomrule
\end{tabular}
\end{table}

\begin{figure*}[!t]
\centering
\includegraphics[width=0.96\linewidth]{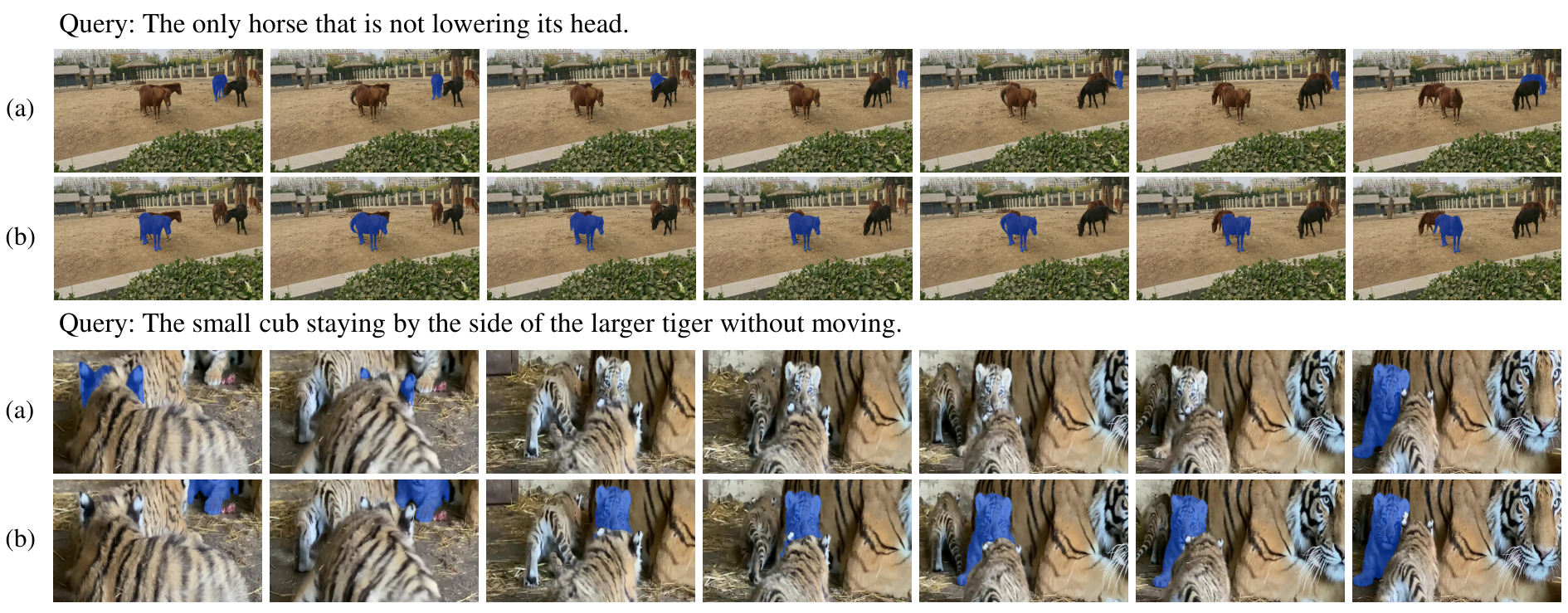}
\caption{Qualitative comparison on MeViS. (a) FlowRVS; (b) SGG. Our predictions preserve the spatial completeness and temporal consistency of the target masks.}
\label{vis_res}
\end{figure*}

\noindent\textbf{Qualitative Results.} \cref{vis_res} presents the masks predicted by FlowRVS and SGG on MeViS. SGG delineates the referred objects more accurately and exhibits fewer inconsistencies between frames.

\subsection{Model Analysis}

To facilitate ablation studies, we use the MeViS valid\_u split, as evaluation on valid requires submissions to the official server.

\noindent\textbf{Components Analysis.} The baseline comprises a DiT generation backbone and a VAE encoder--decoder. As reported in \cref{components}, adding SEG while retaining VAE decoding yields 61.7\% $\mathcal{J}\&\mathcal{F}$, an improvement of 1.5\% over the baseline. When VRD is used alone, it takes the baseline's target-mask latent as input and incorporates SAM2-based decoding, reaching 61.7\% $\mathcal{J}\&\mathcal{F}$ with a gain of 1.5\% over the baseline. Enabling both SEG and VRD gives the full SGG model the highest $\mathcal{J}\&\mathcal{F}$ score of 63.6\%. Together, these results demonstrate the effectiveness of our approach.

\begin{table}[t]
\centering
\footnotesize
\setlength{\tabcolsep}{5pt}
\caption{Component ablation of SGG on MeViS valid\_u.}
\label{components}
\begin{lrbox}{\sggComponentsTableBox}
\begin{tabular}{l | c c | c c c}
\toprule
Method & SEG & VRD & $\mathcal{J}\&\mathcal{F}$ & $\mathcal{J}$ & $\mathcal{F}$ \\
\midrule
Baseline &  &  & 60.2 & 55.7 & 64.6 \\
SGG & \ding{51} &  & 61.7 & 57.5 & 66.0 \\
SGG &  & \ding{51} & 61.7 & 57.5 & 66.0 \\
SGG & \ding{51} & \ding{51} & \textbf{63.6} & \textbf{59.4} & \textbf{67.7} \\
\bottomrule
\end{tabular}
\end{lrbox}
\global\sggAblationTableWidth=\wd\sggComponentsTableBox
\usebox{\sggComponentsTableBox}
\end{table}

\noindent\textbf{Semantically Enhanced Generation.} We examine how SEG's components and cross-expert interaction patterns affect referring video object segmentation performance through the ablations in \cref{sgg-seg-ablation}. In \cref{sgg-seg-ablation}(a), semantic expert interaction (SEI) and Text Modulation (TM) individually improve $\mathcal{J}\&\mathcal{F}$ by 1.1\% and 1.0\% over the configuration without either component. Together, they achieve 63.6\% $\mathcal{J}\&\mathcal{F}$, supporting their complementary contributions. \cref{sgg-seg-ablation}(b) compares K/V reading in the mask expert (ME) and semantic expert (SE), with the semantic expert, text modulation, and auxiliary supervision otherwise unchanged. Self reading (SR) uses only the expert's own K/V, whereas joint reading (JR) uses K/V from both experts. Enabling JR only in ME establishes one-way semantic-to-mask interaction, achieving 63.0\% $\mathcal{J}\&\mathcal{F}$ and improving over independent SR in both experts by 0.3\%. With JR in both experts, bidirectional interaction brings a further gain of 0.6\%, reaching the best $\mathcal{J}\&\mathcal{F}$ of 63.6\%.

\begin{table}[!t]
\centering
\footnotesize
\setlength{\tabcolsep}{5pt}
\caption {Ablation study of SEG configurations on MeViS valid\_u.}
\label{sgg-seg-ablation}
\begin{minipage}[t]{\linewidth}
\centering
\textbf{(a) SEG components.}\par\smallskip
\begin{tabular*}{\sggAblationTableWidth}{@{\hspace{\tabcolsep}\extracolsep{\fill}}l | c c | c c c}
\toprule
Method & SEI & TM & $\mathcal{J}\&\mathcal{F}$ & $\mathcal{J}$ & $\mathcal{F}$ \\
\midrule
SGG &  &  & 61.7 & 57.5 & 66.0 \\
SGG &  & \ding{51} & 62.7 & 58.5 & 66.8 \\
SGG & \ding{51} &  & 62.8 & 58.6 & 66.9 \\
SGG & \ding{51} & \ding{51} & \textbf{63.6} & \textbf{59.4} & \textbf{67.7} \\
\bottomrule
\end{tabular*}
\end{minipage}
\par\medskip
\begin{minipage}[t]{\linewidth}
\centering
\textbf{(b) Cross-expert K/V reading.}\par\smallskip
\begin{tabular*}{\sggAblationTableWidth}{@{\hspace{\tabcolsep}\extracolsep{\fill}}l | c c | c c c}
\toprule
Method & ME & SE & $\mathcal{J}\&\mathcal{F}$ & $\mathcal{J}$ & $\mathcal{F}$ \\
\midrule
SGG & SR & SR & 62.7 & 58.5 & 66.8 \\
SGG & JR & SR & 63.0 & 58.7 & 67.2 \\
SGG & JR & JR & \textbf{63.6} & \textbf{59.4} & \textbf{67.7} \\
\bottomrule
\end{tabular*}
\end{minipage}
\end{table}

\begin{table}[!t]
\centering
\footnotesize
\setlength{\tabcolsep}{5pt}
\caption{Ablation study of VRD configurations on MeViS valid\_u.}
\label{sgg-vrd-ablation}
\begin{tabular}{l | c c c | c c c}
\toprule
Method & DP & LE & SP & $\mathcal{J}\&\mathcal{F}$ & $\mathcal{J}$ & $\mathcal{F}$ \\
\midrule
SGG & \ding{51} &  &  & 62.3 & 58.2 & 66.5 \\
SGG &  &  & \ding{51} & 58.1 & 54.0 & 62.3 \\
SGG & \ding{51} &  & \ding{51} & 62.9 & 58.8 & 67.0 \\
SGG & \ding{51} & \ding{51} &  & 63.1 & 59.0 & 67.2 \\
SGG & \ding{51} & \ding{51} & \ding{51} & \textbf{63.6} & \textbf{59.4} & \textbf{67.7} \\
\bottomrule
\end{tabular}
\end{table}

\noindent\textbf{Visually Refined Decoding.} We investigate how VRD's prompt construction and latent enhancement affect segmentation performance through the ablations in \cref{sgg-vrd-ablation}. We focus on three components: the dense prompt (DP), latent enhancement (LE), and the sparse prompt (SP). DP is derived from the target-mask latent to provide spatial guidance on the target region, while SP is obtained by projecting the aggregated DINOv3 object representation to convey object-level foreground information. All variants use SAM2 and the same SEG checkpoint. We first evaluate DP without latent enhancement and SP separately, followed by their combination. We then enable latent enhancement without a sparse prompt and finally evaluate the complete VRD. LE is evaluated as a unified operation that refines the latent-derived features before dense prompt construction. These comparisons clarify the role of latent enhancement in strengthening spatial guidance and the complementary contributions of spatial and object-level cues to accurate target segmentation.

As shown in \cref{sgg-vrd-ablation}, using DP and SP separately yields $\mathcal{J}\&\mathcal{F}$ scores of 62.3\% and 58.1\%, respectively. Combining them improves over the individual configurations by 0.6\% and 4.8\%, supporting the complementary roles of spatial guidance and object-level visual representations. Without a sparse prompt, adding LE to DP improves $\mathcal{J}\&\mathcal{F}$ by 0.8\%, demonstrating the benefit of enhancing the latent-derived features before decoding. With SP retained, LE provides a further 0.7\% gain over DP + SP. The complete model also exceeds DP + LE by 0.5\% and achieves the best $\mathcal{J}\&\mathcal{F}$ of 63.6\%, indicating that enhanced dense prompts and sparse prompts remain complementary.

\FloatBarrier
\section{Conclusion}
We present SGG, a generative RVOS framework that unifies language-conditioned target generation and visually grounded mask refinement through semantic representations. SEG couples a trainable semantic expert with the mask-generating expert via gated spatiotemporal attention, while text modulation and text cross-attention enhance target-specific modeling within reconstruction-oriented video latents. VRD reformulates latent decoding as SAM2-based prompt-conditioned segmentation, leveraging DINOv3 features to construct a shared object representation that provides sparse prompts and guides spatial refinement for dense prompt construction. These complementary prompts, together with SAM2 video-frame features, enable accurate and complete mask prediction while preserving the referring semantics encoded in the generated target-mask latent. Extensive experiments on multiple RVOS benchmarks demonstrate that SGG achieves state-of-the-art segmentation performance.

\appendix
\clearpage
\twocolumn[
\begin{center}
\Large\bfseries Supplementary Material\\
\end{center}
]

\section{Additional Experimental Studies}

We extend the model analysis in the main paper by examining semantic integration in SEG and the construction of complementary prompts in VRD. All ablation experiments use MeViS valid\_u~\cite{ding2023mevis} and follow the training and evaluation settings of the main paper. The tables report region similarity $\mathcal J$, boundary accuracy $\mathcal F$, and their mean $\mathcal J\&\mathcal F$.

\subsection{Semantically Enhanced Generation}

\noindent\textbf{Semantic Integration.} We investigate whether directly attending to DINOv3 features can replace the trainable semantic expert. Direct cross-attention (DCA) projects frozen DINOv3 features into keys and values for the mask expert, without a semantic expert or its auxiliary prediction objective. Text modulation and the existing text cross-attention are retained. We compare DCA with gated expert interaction (GEI), the design adopted in SEG, using the same decoding procedure for final segmentation. As reported in \cref{tab:supp-seg}(a), DCA obtains 62.4\% $\mathcal J\&\mathcal F$, whereas GEI reaches 63.6\%, improving by 1.2\%. This comparison supports adapting semantic representations through a dedicated expert rather than using them solely as fixed attention context.

\noindent\textbf{Semantic Supervision Weight.} We further examine the balance between flow matching and semantic supervision by varying $\lambda_{\mathrm{sem}}$ over $\{0.5,1,5,10\}$, while keeping $\lambda_{\mathrm{flow}}=1$ and the remaining settings unchanged. \cref{tab:supp-seg}(b) summarizes the results. Setting $\lambda_{\mathrm{sem}}$ to 0.5 or 1 yields 63.2\% and 63.4\% $\mathcal J\&\mathcal F$, respectively. Increasing it to 5 gives the best result of 63.6\%, with gains of 0.4\% and 0.2\% over these two settings. A larger weight of 10 produces 63.3\% and brings no further improvement. We therefore use $\lambda_{\mathrm{sem}}=5$ in the final model.

\begin{figure*}[t]
\centering
\includegraphics[width=0.98\linewidth]{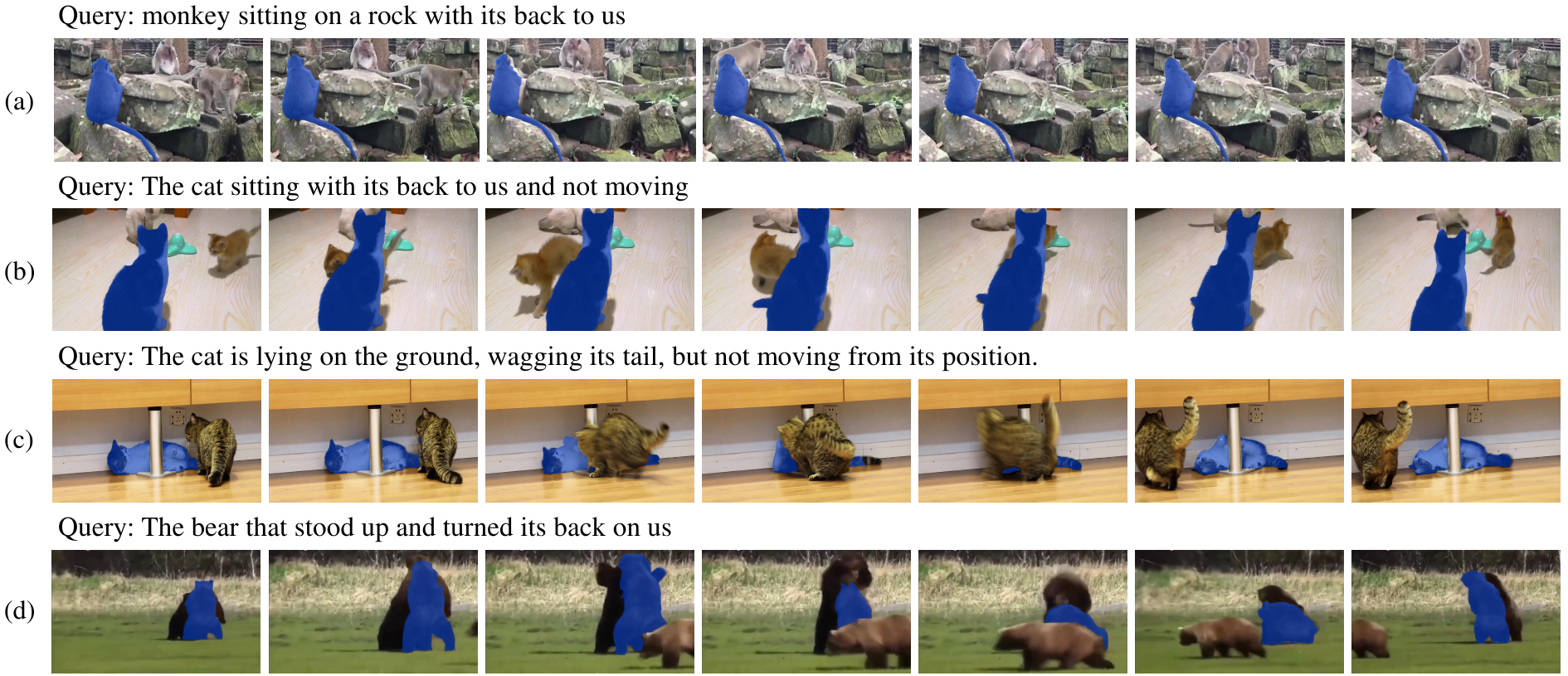}
\caption{Additional qualitative results of SGG on MeViS. Our predictions preserve the target objects' spatial completeness and temporal consistency.}
\label{fig:supp-mevis-examples}
\end{figure*}

\begin{table}[!b]
\centering
\normalsize
\setlength{\tabcolsep}{5pt}
\caption{Analysis of SEG configurations in SGG on MeViS valid\_u.}
\label{tab:supp-seg}
\begin{minipage}[t]{\linewidth}
\centering
\textbf{(a) Semantic integration.}\par\smallskip
\begin{tabular*}{0.82\linewidth}{@{\hspace{\tabcolsep}\extracolsep{\fill}}l | c | c c c}
\toprule
Method & Setting & $\mathcal J\&\mathcal F$ & $\mathcal J$ & $\mathcal F$ \\
\midrule
SGG & DCA & 62.4 & 58.0 & 66.8 \\
SGG & GEI & \textbf{63.6} & \textbf{59.4} & \textbf{67.7} \\
\bottomrule
\end{tabular*}
\end{minipage}
\par\medskip
\begin{minipage}[t]{\linewidth}
\centering
\textbf{(b) Semantic supervision weight.}\par\smallskip
\begin{tabular*}{0.82\linewidth}{@{\hspace{\tabcolsep}\extracolsep{\fill}}l | c | c c c}
\toprule
Method & $\lambda_{\mathrm{sem}}$ & $\mathcal J\&\mathcal F$ & $\mathcal J$ & $\mathcal F$ \\
\midrule
SGG & 0.5 & 63.2 & 58.9 & 67.4 \\
SGG & 1 & 63.4 & 59.3 & 67.5 \\
SGG & 5 & \textbf{63.6} & \textbf{59.4} & \textbf{67.7} \\
SGG & 10 & 63.3 & 59.1 & 67.6 \\
\bottomrule
\end{tabular*}
\end{minipage}
\end{table}

\subsection{Visually Refined Decoding}

\noindent\textbf{Decoding Variants.} We investigate different ways of using DINOv3 features for decoding in VRD. To this end, we design the following two variants. All variants retain Initial local read, Object pooling, and Sparse projection in Fig. 2 of the main paper, while differing in how spatial features condition SAM2 decoding. Direct feature injection (DFI) replaces the dense prompt construction pathway after Initial local read with a feature projection. The projected $H^0$ is integrated into SAM2's memory-conditioned video-frame features through a residual connection, without using SAM2's native dense prompt encoding pathway. Object-guided prompt correction (OPC) retains object-to-frame matching within Object-guided local read but replaces kernel adjustment and the second local read with a convolutional prompt-correction head. The matching relations, original latent-derived features $H$, and unenhanced mask logits predicted by Mask head jointly determine a residual correction to the dense prompt before encoding. Our object-guided local read (OLR) instead uses the matching relations to adjust local aggregation kernels and refine $H$ into $H^1$ before Mask head, while retaining the Object pooling--Sparse projection pathway for sparse prompting. As shown in \cref{tab:supp-vrd}(a), DFI and OPC achieve 63.0\% and 62.6\% $\mathcal J\&\mathcal F$, respectively. OLR yields the best score of 63.6\%, exceeding these alternatives by 0.6\% and 1.0\%. These results support using object-level information to guide feature refinement within prompt-conditioned decoding.

\noindent\textbf{Shared Object Representation.} We examine how the source of the shared object representation affects dense prompt refinement and sparse prompt construction in VRD. A simple variant is latent object pooling (LOP), which spatially averages the target-mask latent, aligns the resulting vectors to the video frames, and projects them into the shared object space. LOP replaces the original Object pooling operation with latent-based pooling. The resulting representation is constructed without DINOv3 features and supplied to both Object-guided local read and Sparse projection. Visual object pooling (VOP) follows the complete VRD design, using spatial weights derived from $H^0$ to aggregate DINOv3 features into the shared object representation. Both configurations retain dense prompt enhancement and sparse prompting; the object representation in each configuration guides local kernel adjustment and is projected into a sparse prompt. In \cref{tab:supp-vrd}(b), LOP achieves 63.1\% $\mathcal J\&\mathcal F$, while VOP, corresponding to the complete model, reaches 63.6\%, a gain of 0.5\%. These results support incorporating DINOv3 semantics into the shared object representation to jointly guide spatial refinement and object-level foreground prompting.

\noindent\textbf{Latent Enhancement.} Starting from the complete VRD, we examine the initial local read (LR) and object-to-frame matching (OM) within dense prompt enhancement while keeping the remaining components unchanged. LR aggregates neighboring DINOv3 features to update the latent-derived spatial features. OM uses the pooled object representation to compute matching relations, adjust the local kernels, and guide a second read. The residual update is applied to the original latent-derived features. All configurations retain the initial local read and Object pooling for DINOv3-derived sparse prompt construction. The LR and OM indicators specify the enhancement used by the dense prompt pathway. The first row of \cref{tab:supp-vrd}(c) uses the unenhanced dense prompt, while the second uses LR alone. Their $\mathcal J\&\mathcal F$ scores are 62.9\% and 63.3\%, respectively, corresponding to a gain of 0.4\%. Combining LR with OM provides a further improvement of 0.3\% and reaches 63.6\% $\mathcal J\&\mathcal F$. This analysis demonstrates how local visual aggregation and object-guided kernel adjustment contribute to the complete refinement process.

\begin{table}[tbp]
\centering
\normalsize
\setlength{\tabcolsep}{5pt}
\caption{Analysis of VRD configurations in SGG on MeViS valid\_u.}
\label{tab:supp-vrd}
\begin{minipage}[t]{\linewidth}
\centering
\textbf{(a) Decoding variants.}\par\smallskip
\begin{tabular*}{0.82\linewidth}{@{\hspace{\tabcolsep}\extracolsep{\fill}}l | c | c c c}
\toprule
Method & Setting & $\mathcal J\&\mathcal F$ & $\mathcal J$ & $\mathcal F$ \\
\midrule
SGG & DFI & 63.0 & 58.8 & 67.1 \\
SGG & OPC & 62.6 & 58.5 & 66.6 \\
SGG & OLR & \textbf{63.6} & \textbf{59.4} & \textbf{67.7} \\
\bottomrule
\end{tabular*}
\end{minipage}
\par\medskip
\begin{minipage}[t]{\linewidth}
\centering
\textbf{(b) Shared object representation.}\par\smallskip
\begin{tabular*}{0.82\linewidth}{@{\hspace{\tabcolsep}\extracolsep{\fill}}l | c | c c c}
\toprule
Method & Pooling & $\mathcal J\&\mathcal F$ & $\mathcal J$ & $\mathcal F$ \\
\midrule
SGG & LOP & 63.1 & 59.0 & 67.3 \\
SGG & VOP & \textbf{63.6} & \textbf{59.4} & \textbf{67.7} \\
\bottomrule
\end{tabular*}
\end{minipage}
\par\medskip
\begin{minipage}[t]{\linewidth}
\centering
\textbf{(c) Latent enhancement.}\par\smallskip
\begin{tabular*}{0.82\linewidth}{@{\hspace{\tabcolsep}\extracolsep{\fill}}l | c c | c c c}
\toprule
Method & LR & OM & $\mathcal J\&\mathcal F$ & $\mathcal J$ & $\mathcal F$ \\
\midrule
SGG &  &  & 62.9 & 58.8 & 67.0 \\
SGG & \ding{51} &  & 63.3 & 59.1 & 67.5 \\
SGG & \ding{51} & \ding{51} & \textbf{63.6} & \textbf{59.4} & \textbf{67.7} \\
\bottomrule
\end{tabular*}
\end{minipage}
\end{table}

\section{Additional Visualization Results}

\noindent\textbf{Additional Results on MeViS.} \cref{fig:supp-mevis-examples} presents additional qualitative results of SGG on MeViS~\cite{ding2023mevis}. The first two examples, \cref{fig:supp-mevis-examples}(a,b), examine target identification using orientation and motion state in scenes containing multiple animals. In the monkey sequence, SGG segments the monkey sitting on a rock with its back to the camera, despite nearby monkeys occupying the same rocky environment. In the cat sequence, the model consistently selects the stationary cat facing away from the camera while another cat moves nearby. These results illustrate the ability to associate the referring expression with the intended object and maintain this association as neighboring objects change position.

The latter two examples, \cref{fig:supp-mevis-examples}(c,d), further examine occlusion and motion. The lying cat moves its tail while remaining in place, and another cat passes in front of it, temporarily obscuring much of its body. SGG retains the visible target regions during occlusion and delineates the target again as it becomes visible, without switching to the moving distractor. The bear sequence combines standing up and turning away with substantial posture changes and partial occlusion by a bear crossing the foreground. SGG follows the referred bear across these changes and adapts the masks to its evolving visible extent. Together, these examples demonstrate spatially coherent segmentation and temporally consistent target association under occlusion and motion.

{
    \small
    \bibliographystyle{ieeenat_fullname}
    \bibliography{main}
}

\end{document}